\documentclass[sigconf, final]{acmart}
\AtBeginDocument{%
  }

\acmConference[SIGIR 2025]{SIGIR 2025}{July 13--18,
  2018}{Padua, Italy}

\usepackage{array}
\usepackage{booktabs}
\usepackage{hyperref}
\usepackage{url}
\usepackage{graphicx}      
\usepackage{subcaption} 
\usepackage{tabularx}
\usepackage{xcolor}
\usepackage{soul}
\usepackage{multirow}
\usepackage{siunitx} 
\usepackage{tcolorbox}
\tcbuselibrary{raster,skins}

\usepackage{tikz}

\newcommand{\emphasize}[1]{``#1''}

\newcommand{\ric}[1]{\textcolor{black}{#1}}
\newcommand{\javier}[1]{\textcolor{black}{#1}}
\definecolor{lightred}{rgb}{0.8, 0.3, 1.0}
\newcommand{\takehiro}[1]{\textcolor{black}{#1}}

\newcommand{\todo}[1]{\textcolor{red}{\textbf{TODO}: [#1]}}

\begin{document}

\title{\ric{Are Generative AI Agents Effective \takehiro{Personalized} Financial Advisors?}}




\author{Takehiro Takayanagi}
\email{takayanagi-takehiro590@g.ecc.u-tokyo.ac.jp}
\affiliation{%
  \institution{The University of Tokyo}
  \city{Tokyo}
  \country{Japan}
}

\author{Kiyoshi Izumi}
\email{izumi@sys.t.u-tokyo.ac.jp}
\affiliation{%
  \institution{The University of Tokyo}
  \city{Tokyo}
  \country{Japan}
}

\author{Javier Sanz-Cruzado}
\email{javier.sanz-cruzadopuig@glasgow.ac.uk}
\affiliation{%
  \institution{University of Glasgow}
  \city{Glasgow}
  \country{United Kingdom}
}
\author{Richard McCreadie}
\email{richard.mccreadie@glasgow.ac.uk}
\affiliation{%
  \institution{University of Glasgow}
  \city{Glasgow}
  \country{United Kingdom}
}
\author{Iadh Ounis}
\email{iadh.ounis@glasgow.ac.uk}
\affiliation{%
  \institution{University of Glasgow}
  \city{Glasgow}
  \country{United Kingdom}
}








\renewcommand{\shortauthors}{Takayanagi et al.}

\begin{abstract}
\looseness -1 Large language model-based agents are becoming increasingly popular as a low-cost mechanism to provide personalized, conversational advice, and have demonstrated impressive capabilities in relatively simple scenarios, such as movie recommendations. But how do these agents perform in complex high-stakes domains, where domain expertise is essential and mistakes carry substantial risk? This paper investigates the effectiveness of LLM-advisors in the finance domain, focusing on three distinct challenges: (1) eliciting user preferences when users themselves may be unsure of their needs, (2) providing personalized guidance for diverse investment preferences, and (3) leveraging advisor personality to build relationships and foster trust. Via a lab-based user study with 64 participants, we show that LLM-advisors often match human advisor performance when eliciting preferences, although they can struggle to resolve conflicting user needs. When providing personalized advice, the LLM was able to positively influence user behavior, but demonstrated clear failure modes. Our results show that accurate preference elicitation is key, otherwise, the LLM-advisor has little impact, or can even direct the investor toward unsuitable assets. More worryingly, users appear insensitive to the quality of advice being given, or worse these can have an inverse relationship. Indeed, users reported a preference for and increased satisfaction as well as emotional trust with LLMs adopting an extroverted persona, even though those agents provided worse advice.
\end{abstract}




\maketitle

\section{Introduction}
\javier{Personalized advice plays a crucial role in our society, particularly in complex and high-stakes domains like healthcare and finance. Advisors and professionals in these fields use their expertise to offer personalized guidance and emotional support to their clients, leveraging people's specific preferences and/or circumstances. However, advisory services are often provided at a high cost, effectively excluding a large portion of the population from this critical advice. In the financial domain, to mitigate this issue, automated decision support systems have been widely studied, with a special focus on investment-related predictions, such as financial asset recommendations~\citep{javier,takayanagi}}.

\begin{figure}[t]
    \centering
    \includegraphics[width=0.9\linewidth]{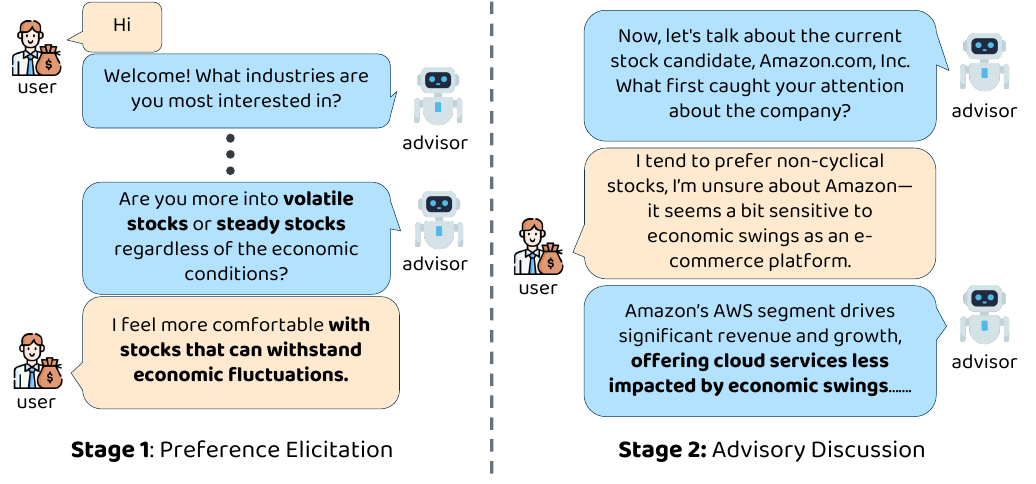}
    \caption{Conceptual illustration of an LLM-advisor with two stages: \takehiro{(1) Preference Elicitation and (2) Advisory Discussion}.}
    \label{fig:conceptual_conversation}
    \vspace{-5mm}
\end{figure}

\takehiro{Recent advances in natural language processing and large language models (LLMs) have significantly accelerated the development of conversational agents, presenting the potential to function as personalized assistants for information-seeking and decision-making~\cite{zamani2023conversational}. These agents \javier{can }now leverage multi-turn dialogues, enabling dynamic, mixed-initiative interactions where both users and systems can take the lead in conversations~\cite{allen1999mixed}. This progression has expanded the application of conversational agents to various tasks, such as recommendation, question answering, and search~\cite{zamani2023conversational,jannach2021survey,sun2018conversational,radlinski2017theoretical}.}

\takehiro{The application of these conversational agents for financial decision-making represents a much more complex scenario than others like movie recommendations, because users are not necessarily familiar with the basic terminology and concepts in this space, and mistakes carry a substantial risk that can lead to large monetary losses.}
\javier{
While there is a growing interest in building these conversational assistants to provide automated financial advice~\cite{lo2024can}, previous work has mostly targeted agents capable of handling simple inquiries~\cite{icaif_finance_advisor,log_analysis_1,log_analysis_2}. Compared to these simple systems, helping users navigate financial decisions and market uncertainties poses a much greater challenge. Therefore, it is not yet clear} how to develop systems that effectively support complex financial information-seeking \javier{and decision-making} tasks. 

\javier{This work aims to close this gap by exploring the effectiveness of LLMs to act as personalized financial advisory agents. In particular, we focus on three problems: (a) eliciting investor preferences through interactive conversations, (b) \takehiro{providing personalized guidance to help users determine whether particular financial assets align with their preferences}, and (c) leveraging the personality of the advisor to foster trust on the advisor.}



\looseness -1 \javier{First, the financial literature emphasizes that eliciting user preferences is central to delivering suitable advice~\cite{streich2023risk}. However, it remains unclear whether current conversational technologies, particularly those powered by LLMs, can correctly elicit user preferences in specialized domains where users struggle to articulate their needs. \takehiro{Our work addresses this challenge in the context of financial services.}} 


Second, although personalization is widely regarded as important in the financial decision-support literature~\cite{takayanagi,javier}, its value in a conversational setting remains uncertain. In particular, \javier{we explore whether tailoring dialogue around a user’s profile and context improves financial decision-making. Additionally, we also explore how personalization influences user perceptions of the advisor, in terms of aspects like trust and satisfaction.}

\takehiro{Finally, in personalized advisory settings within high-stakes domains, the relationship and trust between the client and advisor play a crucial role~\cite{lo2024can}. Research on conversational agents suggests that agent personality significantly affects users’ perceptions of the system~\cite{smestad2019chatbot,cai2022impacts}. However, it remains unclear how an advisor’s personality in the financial domain influences both the quality of users’ financial decisions and their overall experience.}


\javier{To summarize, in this paper, we explore the following questions:}
\begin{itemize}
    \item \textbf{RQ1:} \takehiro{Can LLM-advisors effectively elicit user preferences through conversation?}
    \item \textbf{RQ2:} Does personalization lead to better investment decisions and a more positive advisor assessment?
    \item \textbf{RQ3:} Do different personality traits affect decision quality and advisor assessment?
\end{itemize}



To address these \javier{questions}, we \javier{conduct} a \takehiro{lab-based} user study that \javier{explores the effectiveness of LLMs as interactive conversational financial advisors, on which we simulate realistic investment scenarios using investor narratives and stock relevance scores curated by financial experts.} \javier{Figure~\ref{fig:conceptual_conversation} illustrates an example conversation with the advisor, divided into two stages: first, the LLM-advisor attempts to capture the investor preferences through conversation; in the second \takehiro{stage}, given an individual asset, the advisor provides information about it to the investor, including how the asset matches (or not) the investor's preferences.}
\javier{To answer the different questions, we compare different configurations of the LLM-advisor:} first, we compare personalized vs. non-personalized advisors, and, then, we compare two personalized advisors with \takehiro{distinct} personalities.

\section{Related Work}
\label{s:related_work}
\subsection{Personalization and Preference Elicitation}
\javier{Information systems, especially those focused on search and recommendation benefit from personalization~\cite {komiak2006effects}. Specifically,} personalization techniques play a crucial role in enhancing user experience~\cite{li2016does, pu2011user, zanker2019measuring}. \javier{Interactive approaches, such as conversational preference elicitation represent the frontier of personalization.} This problem has received growing attention, as advances in \ric{generative AI now provide a functional mechanism to collect user preferences dynamically in a free-form manner~\cite{zamani2023conversational}}. This interactive approach \ric{can capture more diverse and targeted insights than static approaches like questionnaires}~\cite{radlinski2019coached,jannach2021survey,radlinski2017theoretical,christakopoulou2016towards,sun2018conversational}. \javier{Indeed, r}ecent studies have proposed various methods for effective conversational preference elicitation~\cite{sun2018conversational,zhang2018towards}, as well as user studies on the perceived quality of this process in domains such as e-commerce, movies, fashion, books, travel, and restaurant recommendations~\cite{kostric2021soliciting,sun2018conversational,radlinski2019coached,argal2018intelligent,de2017recognizing,ziegfeld2025effect}.

\looseness -1 \ric{
However, we argue that for some important domains, trying to directly collect preferences is insufficient. 
An implicit assumption of \takehiro{these studies} is that if directly asked, the user will be able to accurately express their preferences. It is reasonable to expect that this assumption would hold for scenarios like movie recommendation; we can ask a user \emphasize{do you like horror movies?} and expect a useful response. \takehiro{On the other hand}, this will not hold for complex tasks, where the user lacks the knowledge to form an accurate response~\cite{warnestaal2005user,jannach2021survey}. For instance, in an investment context if we asked \emphasize{do you prefer ETFs or Bonds?}, it is not clear that an inexperienced user would be able to produce a meaningful answer. In these cases, an ideal agent needs to fill the gaps in the user knowledge through conversation, as well as infer the user preferences across multiple (often uncertain) user responses. 
But how effective are generative AI agents at this complex task? This paper aims to answer that question for the domain of financial advisory; a particularly challenging domain given its technical nature and high risks if done poorly.}  

\begin{figure*}
\includegraphics[width=1.8\columnwidth]{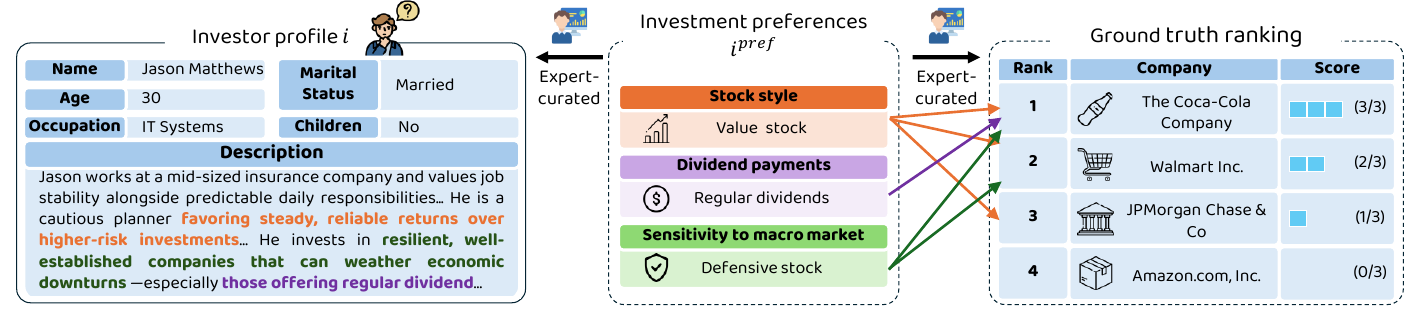}
\caption{\javier{Example of an investor profile, investment preferences, and ground truth ranking. Dashed line components are used for evaluation (and therefore, they are not shown to the user/LLM).} }
\vspace{-4mm}
\label{fig:narrative}
\end{figure*}
\subsection{Financial advisory}
\label{ss:related_work_financial_advisory}
\looseness -1 \takehiro{In the financial \ric{domain, advisors} help individuals manage their personal finances by offering guidance on investments and assisting with decision-making. While financial advisors can be beneficial, their services often come at a high cost, making them unaffordable for many people. To mitigate this issue, automated \ric{(non-conversational)} financial decision support systems such as financial recommender systems have been widely studied~\cite{zibriczky122016recommender}. The majority of research in this area has been focused on how to find profitable assets (i.e. those that will make money if we invest in them). These works assume a simplified user-model, where an investor is only concerned with maximizing return-on-investment over a fixed period of time~\cite{javier,takayanagi}.
These studies frame financial advisory as a ranking problem, where the goal is to rank financial assets for a user over a specified time period.
However, a recent study suggests that a large part of the value offered by human financial advisors stems from their ability to personalize investment guidance to clients’ specific needs, build relationships, and foster trust~\cite{vanguard2020}, rather than simply presenting suitable assets.} 

\javier{Reflecting on these findings, the development of conversational financial advisors has drawn increasing attention, as it enables a dynamic understanding of users’ needs, personalized guidance, and the potential to build trustworthy relationships~\cite{zhao2024revolutionizing,icaif_finance_advisor,hci_finance,hci_2,hildebrand2021conversational}. 
In particular, the conversational agents' personality has gained attention as a factor that can help build relationships with clients and foster trust~\cite{lo2024can}, especially given the successes of conversational agents using the Big Five personality model~\cite{McCrae1992AnIT} to enhance the end-user experience~\cite{CHARNESS201343,streich2023risk}.
Although conversational agents show potential in finance, how to configure them to match the value of human advisors remains unclear. Therefore, we conduct a user study to examine how personalizing investment guidance and the advisor’s personality shape users’ financial decision-making effectiveness and overall user experience.}

\section{Methodology}\label{sec:methodology}

\ric{In this paper we aim to determine to what extent current generative language models can act as an effective financial advisor. Indeed, given the need to personalize for the user, emotional implications, the technical nature of the information-seeking task, and high impact if failed, we argue that this is an excellent test case for the limits of generative large language models. To structure our evaluation, we divide our study into two phases, as illustrated in Figure~\ref{fig:conceptual_conversation}, where we evaluate the success of both:}
\begin{enumerate}
    \item \textbf{Preference Elicitation}: \ric{During this stage, we have the LLM-advisor hold a natural language conversation with a human, where it is directed to collect information regarding the person's investment preferences. The human in this interaction is pretending to have preferences from a given investor profile.}
    \item \textbf{Advisory Discussion}: \ric{During the advisory discussion, the LLM-advisor again has a natural language conversation with the human (acting on an investor profile), where the human collects information about whether a company is a suitable investment for them. This is repeated for multiple companies per investor profile.} 
\end{enumerate}
\looseness -1 \ric{We provide preparatory information and discuss each stage in more detail below:}

\subsection{Investor Profiles}
\label{sec:dataset_investor_profile}

\ric{To fairly evaluate the ability of any LLM-advisor, we need to have them interact with human users with real needs. Given the open-ended nature of free-form conversations, it is desirable to repeat each experiment with different people such that we can observe variances in conversation paths, as those variances may influence task success. However, to enable repeatability, we need to hold the investor needs constant across repetitions. Hence, we define three archetypal investor profiles $i \in I$ based on input from a financial expert, where our human participants are given one to follow when conversing with the LLM-advisor:}

\begin{itemize}
    \item \takehiro{\textbf{\javier{Investor} 1: Growth-Oriented Healthcare Enthusiast:} Prefers healthcare innovations, values high-growth opportunities, and takes measured risks.}
    \item \takehiro{\textbf{\javier{Investor} 2: Conservative Income Seeker:} Seeks stable returns, invests in well-established companies, values regular dividend payouts.}
    \item \takehiro{\textbf{\javier{Investor} 3: Risk-taking Value Investor:} Targets undervalued companies with strong long-term potential, tolerates short-term volatility, and invests in cyclical sectors.}
\end{itemize}

\looseness -1 \javier{For each of} these investor profiles\javier{, we select three key investment preferences,} chosen from well-known investment characteristics such as industry sector, stock style, consistency in dividend payments, and sensitivity to global market changes~\cite{fama1998value}. \javier{We denote the set of investor preferences as $i^{pref}$.} \javier{In our experiments, we simulate a realistic elicitation scenario where the advisor collects the preferences from the participants. Therefore, we do not straightforwardly provide the preferences to the participants. Instead, we present them as} text narratives of between 150 to 200 words. A financial expert was consulted to confirm the quality and reliability of these narratives. An example \javier{narrative representing Investor 2} is illustrated in \javier{Figure~\ref{fig:narrative}, where we highlight the sentences referring to specific investor preferences.}



%

\subsection{Stage 1: Preference Elicitation}
\ric{The goal of stage 1 of our study is to determine to what extent an LLM-advisor can effectively collect a user’s investment preferences through conversation.}
\takehiro{Formally, \javier{given a participant of the user study $u$ and an investor profile $i$, during the elicitation stage, the LLM-advisor aims to obtain an approximated set of preferences, denoted $i_u^{LLM}$, that matches the investor preferences ($i^{pref}$). To achieve this,}}
\takehiro{the generative model}
\takehiro{produce\javier{s} a series of questions \javier{ that participants answer by interpreting the investor narrative. Responses to those questions, denoted as $R^u_i$, are used by the LLM-advisor to generate the user profile $i^{LLM}_u$. }}
\ric{Success is then measured by manually evaluating the overlap between $i^{pref}$ and $i_u^{LLM}$.}




\takehiro{For user elicitation, we adopted a System-Ask-User-Respond (SAUR) paradigm~\cite{zhang2018towards}. During the conversation, the advisor proactively inquires about the user’s preferences given a set of target preferences} 
\takehiro{(e.g.,industry type, acceptable risk). After the \javier{human participant} responds to a question, the LLM-advisor checks whether the \javier{collected} preferences cover all of the target preferences. If the advisor is confident that they do, it ends the conversation and prompts the user to proceed to the next \javier{stage}; otherwise, it continues asking follow-up questions in a loop.}

\subsection{Stage 2: Advisory Discussion}\label{ss:ad}

\ric{Stage 2 of our study investigates to what extent an LLM-advisor can provide the same benefits as a real human advisor when exploring investment options. Note that the goal here is not to have the LLM-advisor promote any one asset, but rather to provide accurate and meaningful information such that the human can find the best investment opportunity for them. To this end, we structure our experiment such that the human (acting on an investor profile) has one conversation with the LLM-advisor for each of a set of assets being considered.\footnote{These were manually selected, however in a production environment these might be produced by an asset recommendation system.} \javier{After all assets are presented to the participant, a stock ranking is generated by sorting the stocks by the participant rating in descending order.}} 

\ric{Importantly, as we know the investor profile \javier{$i^{pref}$} for each conversation about an asset $a$, we can objectively determine whether $a$ is a good investment given \javier{$i^{pref}$},}  
\ric{forming a ground truth against which we can compare to the rating provided by our human participant after their conversation with the LLM-advisor. \javier{For each asset $a$, a financial expert produced a score between 0 and 3 } by manually checking whether $a$ satisfied each of \javier{the} three investment criterial contained in $i^{pref}$. 
\javier{A ground-truth ranking was produced by sorting the assets by the expert scores. We show an example of the ranking construction in Figure~\ref{fig:narrative}.} \javier{During evaluation,} the closer the participant \javier{ranking} is to the \javier{ranking produced by} expert judgment\javier{s}, the better the LLM-advisor performed.}

\enlargethispage{\baselineskip} 
\vspace{2mm}\noindent \textbf{Baseline Prompt}: \ric{As we are working with an \takehiro{LLM-advisor} and the nature of financial information-seeking is time-sensitive, we need to provide any information that might change over time to the LLM within the prompt. As such, for each asset $a$, we pre-prepared a standard asset descriptor block after consulting with a financial expert, containing:}

\begin{itemize}
    \item \textbf{Stock Prices}: We collect monthly stock prices from 2023 using 
    \href{https://finance.yahoo.com/}{Yahoo! Finance}.\footnote{The scenario for the financial advising of our user study is set to December 30, 2023. By basing our experiment at the end of 2023, we avoid the problem of data contamination~\cite{sainz-etal-2023-nlp}.}

    \item \textbf{Business Summary}: We gather each company's business overview from 
    \href{https://finance.yahoo.com/}{Yahoo! Finance}.

    \item \textbf{Recent Performance and Key Financial Indicators (e.g., EPS)}: We obtain earnings conference call transcripts\footnote{Earnings conference calls, hosted by publicly traded companies, discuss key aspects of their earnings reports and future goals with financial analysts and investors, thus covering critical financial indicators and recent performance insights~\cite{sourav2022}. These transcripts cover significant financial indicators and provide explanations of recent performance.} from 
    \href{https://seekingalpha.com/}{Seeking Alpha} for the last quarter of 2023.
\end{itemize}
\looseness -1 \ric{The advisor using this prompt acts as our baseline for the advisory discussion study. We augment this baseline with additional context and instructions to form two additional experimental scenarios, discussed below:}

\vspace{2mm}\noindent \textbf{+Personalization}: \ric{As discussed earlier, one of the core roles of the financial advisor is to personalize to the individual customer, based on their financial situation, needs, and preferences. To enable the LLM-advisor to personalize for the user, we integrate the generated profile from the preference elicitation (Stage 1) $i_u^{LLM}$ into the prompt. \javier{We represent each preference as a series of short sentences.}} 

\vspace{2mm}\noindent \textbf{+Personality}: \ric{In Section~\ref{ss:related_work_financial_advisory} we discussed how human financial advisors provide emotional support as well as financial advice. While it is unlikely that an LLM-advisor could do this as well as a human (it lacks both emotional intelligence and non-conversational clues to the customer's mental state~\cite{wang2023emotional}) it might be possible to provide a better end-user experience by directing the LLM-advisor to adopt a \emph{personality}. As noted in Section~\ref{s:related_work} it is possible to do this via prompt engineering, such as instructing the LLM to take on the traits of one or more of the Big-Five personality types~\cite{McCrae1992AnIT}. }

\ric{As we are performing a user study with humans, it would be impractical to exhaustively test every combination of personality types, hence as an initial investigation}
\takehiro{we experiment with two distinct personality profiles~\cite{smestad2019chatbot}:}
\begin{itemize}
\item \textbf{Extroverted}: \takehiro{High in extroversion, agreeableness, and openness; low in conscientiousness and neuroticism.}
\item \textbf{Conscientious}: \takehiro{Low in extroversion, agreeableness, and openness; high in conscientiousness and neuroticism.}
\end{itemize}

We adopted the prompting method from Jiang et al. (2024) to assign a Big Five personality trait to the LLM agent~\cite{jiang-etal-2024-personallm}, choosing it for its simplicity and effectiveness among various proposed approaches for embedding personality in LLMs (including both prompting and fine-tuning)~\cite{jiang2024evaluating,jiang-etal-2024-personallm,shao-etal-2023-character}. To ensure a high standard of professionalism and accurate representation of the intended personality, we consulted financial professionals to review the texts generated by \takehiro{LLMs adopting both personas.}

\label{sec:user_study_design}
\begin{figure}
    \centering
    \includegraphics[width=0.95\linewidth]{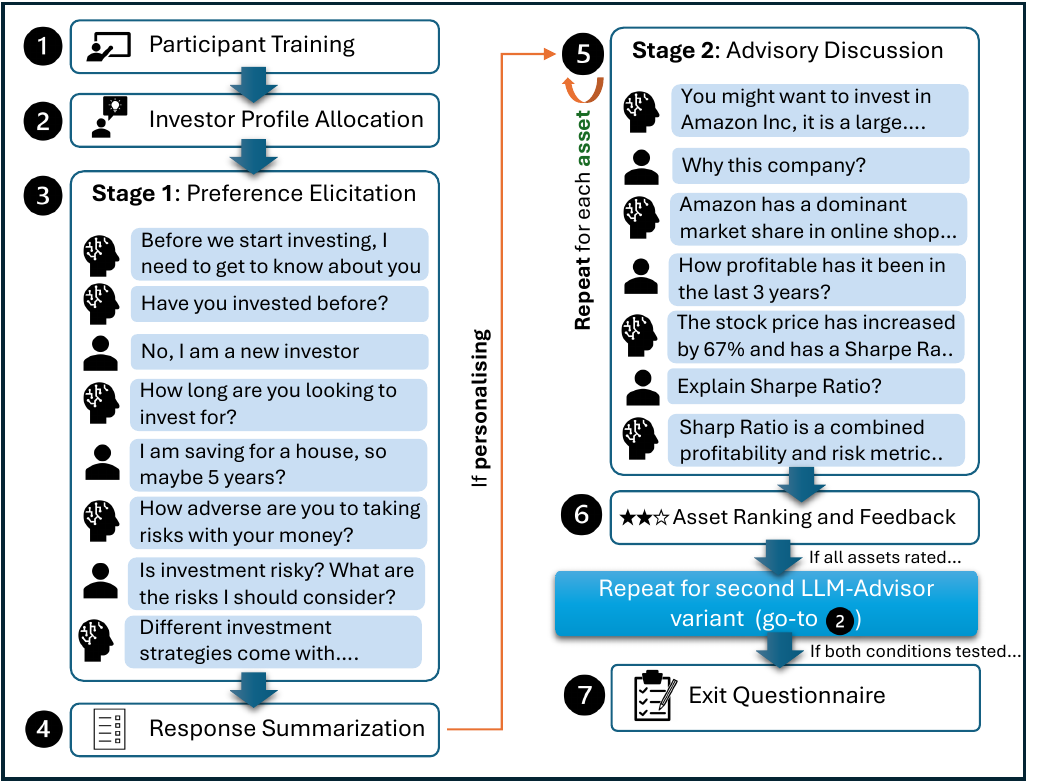}
    \caption{User study structure.}
    \vspace{-5mm}
    \label{fig:user_study_design}
\end{figure}

\subsection{Experimental Design}

\takehiro{In our experiment, we conducted two studies: a personalization study (for RQ2) and an advisor persona study (for RQ3). In the personalization study, participants compared a non-personalized (Baseline) advisor with a personalized (+Personalized) version. In the advisor persona study, they compared different LLM-advisor personality types (+Extroverted vs. +Conscientious). Participants are randomly assigned to one of these two studies.}

Figure~\ref{fig:user_study_design} shows the structure of our user study for a single participant, comprising seven steps:
\begin{enumerate}
    \item \textbf{Participant Training}: \takehiro{Participants are given a general overview of the user study and given instructions on their expected roles during preference elicitation, advisory discussions, asset ranking, and advisor assessment.}
    \item \textbf{Investor Profile Allocation}: The user $u$ is randomly allocated one of the investor profiles (See Section~\ref{sec:dataset_investor_profile}) that they will follow. Each profile is assigned to \takehiro{42} 
    participants.
    \item \textbf{Preference Elicitation (Stage 1)}: The participant interacts with the LLM-advisor as if they were a new investor. The conversation ends once the LLM-advisor determines that they know enough about the investor to personalize for them. \takehiro{The median time spent on preference elicitation was 5 minutes and 11 seconds.} 
    \item \textbf{Response Summarization}: 
    \takehiro{Given the aggregator of user responses $R^u_i$, we instruct an LLM to generate a\javier{n investor} profile $i^{LLM}_u$. For \javier{each investor preference in $i^{pref}$}, if there is any relevant information in the responses $R^u_i$, that information is included in $i^{LLM}_u$. Otherwise, $i^{LLM}_u$ indicates that no relevant information is available \javier{for that specific preference}.}

    \item \textbf{Advisory Discussion (Stage 2)}: \ric{To simplify the conversation flow we have the participant hold separate conversations with the LLM-advisor for each asset they might invest in. The LLM-advisor is provided with context about the current asset (see Section~\ref{ss:ad}), and depending on the experimental scenario, optionally personalization information (step 4 output) and/or a target personality context statement. Each conversation continues until the user is satisfied that they have enough information to rate the asset.} \takehiro{The order in which the assets are discussed is randomly assigned to avoid position bias.} 
    \item \textbf{Asset Ranking \takehiro{and Feedback}}: \takehiro{Participants rank all the stocks (four in total) discussed in the advisory session according to their desire to invest in each. They also assess the advisor they interacted with using a 7-point Likert scale for the items listed in Table~\ref{tab:operational_definition} (see Section~\ref{ss:metrics}).}
\end{enumerate}

\noindent \looseness -1 \ric{To enable more effective pair-wise comparison of LLM-advisor variants, we have each participant test \emph{two variants per \takehiro{study}}. 
If the user has only tested one variant at this point, then they repeat the user study (starting at step \takehiro{2}) with the second variant.} \takehiro{The order in which participants experience each variant is randomly assigned.}

\begin{enumerate}
    \setcounter{enumi}{6}
    \item \textbf{Exit Questionnaire}: \ric{Once a pair of LLM-advisor variants have been tested, the user fills in an exit questionnaire that is designed to \takehiro{ask the overall experience in the user study.}}
\end{enumerate}

\javier{In our experiments, we use Llama-3.1 8B as the background model for all our LLM-advisor variants.\footnote{Further details about the LLM configuration, investor narratives, relevant scores, prompts and scripts for data analysis can be accessed at the 
following repository: \href{https://github.com/TTsamurai/LLMAdvisor_supplementary}{https://github.com/TTsamurai/LLMAdvisor\_supplementary}}}

\subsection{Participants}
\looseness -1 We recruited 64 participants from the authors' affiliated university for our study: 32 participants for the \takehiro{personalization study} and 32 participants for the \takehiro{advisor persona} study, utilizing the university's online platform and blackboard for recruitment. Participants were required to be fluent in English, over 18 years old, and have an interest in finance and investment, mirroring the target demographic of our system's users. After excluding invalid data, 29 participants remained in the \takehiro{personalization} study and 31 in the \takehiro{advisor persona study}. We conducted a power analysis using the Wilcoxon signed-rank test for matched pairs, with the experimental conditions as the independent variable and users' response to the advisor assessment questionnaire as the dependent variable~\cite{sakai2018laboratory}. The analysis determined that 29 participants are needed to observe a statistically significant effect on user-perceived quality. Our recruitment criteria and compensation \ric{(£10/hour)} for approximately one hour of participation were approved by our organization's ethical board.

\newcolumntype{M}[1]{>{\centering\arraybackslash}m{#1}}
\newcolumntype{L}[1]{>{\raggedright\arraybackslash}m{#1}}
\begin{table}[t]
\centering
\caption{\takehiro{Operational definitions used in the advisor assessment questionnaire for all response dimensions.}}
\small 
\resizebox{1.0\linewidth}{!}{
\begin{tabular}{L{4cm}L{7cm}}
\toprule
\textbf{Response Dimension} & \textbf{Operational Definition} \\ 
\midrule
Perceived Personalization~\cite{komiak2006effects} & The advisor understands my needs. \\ 
\midrule[0.05pt]
Emotional Trust~\cite{komiak2006effects} & I feel content about relying on this advisor for my decisions. \\ 
\midrule[0.05pt]
Trust in Competence~\cite{komiak2006effects} & The advisor has good knowledge of the stock. \\ 
\midrule[0.05pt]
Intention to Use~\cite{komiak2006effects} & I am willing to use this advisor as an aid to help with my decision about which stock to purchase. \\ 
\midrule[0.05pt]
Perceived Usefulness~\cite{pu2011user} & The advisor gave me good suggestions. \\ 
\midrule[0.05pt]
Overall Satisfaction~\cite{pu2011user} & Overall, I am satisfied with the advisor. \\ 
\midrule[0.05pt]
Information Provision~\cite{vanichvasin2021chatbot} & The advisor provides the financial knowledge needed. \\ 
\bottomrule
\end{tabular}}
\label{tab:operational_definition}
\vspace{-3mm}
\end{table}


\section{Evaluation Metrics and Statistics}\label{ss:metrics}

\ric{In this section we discuss how we quantify effectiveness for the preference elicitation and advisory discussion stages, respectively, in addition to summarizing dataset statistics for each.} 

\subsection{Preference Elicitation Metrics (Stage 1)}\label{ss:stage1}
\ric{To evaluate the quality of the first preference elicitation stage, we want to measure how well the LLM-advisor has captured the investor preferences as defined in the investor profile $i$ (see Section~\ref{sec:dataset_investor_profile}). Each investor profile $i \in I$ defines key features of the investor, such as preferring high-growth stocks, or favoring regular payouts, denoted $i^{pref}$. We have three investor profiles ($|I|$=3), with \takehiro{10} ($n$) participants performing elicitation on \takehiro{$i^{LLM}_u$ for each profile and each LLM variant, i.e. there are 120 elicitation attempts in total, with 30 attempts per LLM‐advisor variant.} \javier{Following the notation in Section~\ref{sec:methodology},} $i^{LLM}_u$ in this case denotes a similar list of features to $i^{pref}$ that LLM-advisor learned about the investor during conversation \javier{with a participant $u$}, which we derive from a manual analysis of the elicitation output (i.e. what is produced by response summarization). Intuitively, the closer the features produced from any elicitation attempt $i^{LLM}_u$ is to $i^{pref}$, the better the LLM-advisor is performing. To this end, we report elicitation accuracy for each investor profile, calculated as:}
\begin{equation}
    \text{ElicitationAccuracy}(i) = \frac{1}{n} \sum_{j=1}^n \frac{\left|i_j^{LLM} \cap i^{pref}\right|}{\left|i^{pref}\right|}
\end{equation}



\vspace{2mm}\noindent  \ric{\textbf{Human Advisor}: To provide a point of comparison, we also conduct a preference elicitation with a financial expert using the same prompt and instructions as the LLM. This allows us to evaluate how close LLMs are to a paid human advisor undertaking the same task. More specifically, for each investor profile, three participants engaged with this expert, who then produced a set of preferences $i_u^{Expert}$, which can be used instead of $i_u^{LLM}$ in \takehiro{Equation 1.}}

\begin{table}[t]
\centering
\small
\looseness -1 \caption{General statistics of the collected conversation data.}
\label{tab:general_stats}
\begin{tabular}{l r}
\toprule
Participants & 60 \\
Time Period & 2024/10/24 \textasciitilde{} 2024/11/7 \\
Total Turns & 10,008 \\
\midrule
\textbf{Stage 1: Preference Elicitation} & \\
\quad Total Turns & 1,788 \\
\quad Number of Sessions & 120 \\
\quad Avg. Turns/Session & 15.8 \\
\quad Avg. User Words/Turn  & 9.8 \\
\midrule
\textbf{Stage 2: Advisory Discussion} & \\
\quad Total Turns & 8,220 \\
\quad Number of Sessions & 480 \\
\quad Avg. Turns/Session & 18.2 \\
\quad Avg. User Words/Turn  & 13.0 \\
\bottomrule
\end{tabular}
\vspace{-3mm}
\end{table}

\subsection{Advisory Effectiveness Metrics (Stage 2)}\label{ss:stage2}

\looseness -1 \ric{\textbf{Ranking correlation (Spearman’s Rho)}: In the second stage, we evaluate how well the LLM-advisor can support an investor to select financial assets that are suitable for them to invest in. Recall from Figure~\ref{fig:user_study_design} that after a participant finishes discussing all assets with the LLM-advisor, they rank those assets $a \in A_i$ based on the likelihood they will invest in each, i.e. each participant $u$ acting on a profile $i$ we have an asset ranking $R(A_i, i_u)$. As illustrated in Figure~\ref{fig:narrative}, each investor profile $i$ was derived from a ground truth set of investor preferences $i^{pref}$, which an expert used to create a ground truth ranking $R(A_i, i^{pref})$, i.e. the \emphasize{correct} ranking of assets. Intuitively the closer the $R(A_i, i_u)$ is to $R(A_i, i^{pref})$, the better the advisor is performing, as the participant was better able to distinguish suitable assets vs. unsuitable ones. Hence, to evaluate the effectiveness of the advisory task, we report the mean ranking correlation (Spearman’s Rho) between $R(A_i, i_u)$ and $R(A_i, i^{pref})$ across participants $u$ for each LLM-advisor.}       


\vspace{2mm} \noindent \textbf{\takehiro{Advisor Assessment Questionnaire}}:
\ric{Lastly, we also gather qualitative data from each participant via a questionnaire. In particular, after ranking assets each participant, reports how they feel the LLM-advisor performed in terms of 7 dimensions, listed in Table~\ref{tab:operational_definition}, such as perceived usefulness, trust, and user satisfaction. We use this data later to evaluate how sensitive the user is to differences in the LLM-advisor.}

\subsection{Dataset Statistics}
\looseness -1 Table~\ref{tab:general_stats} \ric{summarizes the statistics of the data collected during the two stages of our user study. Each conversation that a participant had with an LLM-advisor in either stage 1 or 2 is referred to as a session, e.g. during Stage 1, there were 3 investor profiles * 10 participants * 4 LLM-advisors, resulting in 120 sessions. Stage 2 has 4x the number of sessions, as there are four assets associated with each profile ($A_i$) to discuss with the LLM-advisor. } 

\ric{From Table~\ref{tab:general_stats} we observe that in contrast to other conversational tasks}~\cite{log_analysis_1,log_analysis_2}, financial information-seeking appears to require more extended interactions. On average, preference elicitation involves 15 turns per session with 9.8 words per turn, whereas advisory discussions involve 18 turns per session with 13.0 words per turn, highlighting the overall complexity of the task.

\section{Results}

\looseness -1 In this work, we explore how to design conversational financial advisors that enhance both decision-making and positive experience. To achieve this, our user study is guided by 3 core research questions.

\begin{itemize}
    \item \textbf{RQ1:} Can LLM-advisors effectively elicit user preferences through conversation?
    \item \textbf{RQ2:} Does personalization lead to better decisions and more positive advisor assessment?
    \item \textbf{RQ3:} Do different personality traits affect decision quality and advisor assessment?
\end{itemize}

\begin{table}[t]
\centering
\caption{Stage 1 - Comparison of Elicitation Accuracy of an expert vs. different LLM-advisors for each investor profile. The best advisor is highlighted in bold. Arrows denote percentage increases ($\uparrow$) or decreases ($\downarrow$) compared to the expert.}
\label{tab:preference_elicitation_performance}
\resizebox{\linewidth}{!}{
\begin{tabular}{l c c c c c}
\toprule
\multirow{2}{*}{\textbf{Investor Profile}} 
  & \multirow{2}{*}{\textbf{Expert}} & \multicolumn{4}{c}{LLM-Advisors} \\
   \cmidrule(lr){3-6}
  & & \textbf{LLM} & \textbf{+Extr.} & \textbf{+Cons.} & \textbf{Average} \\
\midrule
\textbf{Growth-Oriented} 
  & 0.78
  & 0.76 
  & \textbf{0.80}
  & 0.79  
  & 0.78$^{\rightarrow0.0\%}$ \\ 
\textbf{Conservative-Income} 
  & \textbf{0.89} 
  & 0.82  
  & 0.75
  & 0.87  
  & 0.82$^{\downarrow7.8\%}$  \\
\textbf{Risk-Taking} 
  & \textbf{0.89} 
  & 0.48  
  & 0.60  
  & 0.55
  & 0.53$^{\downarrow40.5\%}$ \\
\midrule
\textbf{Average} 
  & \textbf{0.85} 
  & 0.69  
  & 0.70
  & 0.73
  & 0.70$^{\downarrow17.6\%}$ \\
\bottomrule
\end{tabular}}
\vspace{-5mm}
\end{table}

\subsection{RQ1: Elicitation accuracy} \label{sec:rq1}

\looseness -1 \ric{We begin by examining how effective the LLM-advisors are at identifying investment preferences during conversations in Stage 1. Elicitation Accuracy is the primary metric, where we contrast the mean accuracy across 10 sessions in comparison to a human expert tackling the same task (see Section~\ref{ss:stage1}). Table~\ref{tab:preference_elicitation_performance} reports elicitation accuracy for each LLM-advisor and the Human Expert across investment profiles. Arrows denote percentage increases ($\uparrow$) or decreases ($\downarrow$) of the LLM-advisor compared to the expert.}

\ric{To set expectations, we first consider the performance of the expert in the first column in Table~\ref{tab:preference_elicitation_performance}, as we might expect, the expert} maintains consistently high performance across all profiles, \ric{averaging 85\% accuracy (random accuracy is 50\%). This forms an expectation of the performance ceiling for the task.}

\ric{Next, we compare the expert performance to each LLM-advisor. From the perspective of preference elicitation, there are three LLM-advisor configurations, those that use only the Baseline Prompt (denoted LLM) \takehiro{from the personalization study}, and those that include a defined personality (either extroverted, +Extr., or conscientious, +Cons.) \takehiro{from the advisor persona study}.}\footnote{Note we cannot have a personalized variant here, as the personalization evidence is derived from this stage.} \ric{From Table~\ref{tab:preference_elicitation_performance}, we observe that the LLM-advisor’s performance is generally strong for growth-oriented, and conservative-income investors (with accuracy around 80\%) on average, which is similar to the human advisor. However, for the risk-taking investor profile, the LLM-advisor’s elicitation accuracy was substantially lower (-40.5\%).}

\ric{From a manual failure analysis, we observed the following trends that contribute to the performance gap with the human advisor, particularly for the risk-taking profile. First, it is notable that elicitation failures can originate from the investor (participant) rather than the LLM. Recall that one of the aspects that makes finance more challenging than domains like movie recommendation is that the \emphasize{user} is inexpert, and so may give incorrect information during the conversation. Indeed, we observed cases where the participant confused concepts such as the difference between a growth and a value stock, as well as cyclical/non-cyclical assets. On the other side, preference hallucination is a core issue for the LLM-advisor. The LLM is a probabilistic token generator conditioned on the baseline prompt and prior conversation, and as a result, in some scenarios, the contextual content can override a statement by the investor. This type of error is more likely when the investor is unsure in their responses or when they provide contradictory statements.}  For instance, an investor expressing an interest in the consumer discretionary sector while simultaneously opting for non-cyclical stocks, despite consumer discretionary being inherently cyclical.

\vspace{2mm} \ric{\textbf{To answer RQ1}, our results demonstrate that LLM-advisor's are able to elicit preferences from a user via conversation and that for 2/3's of the user profiles tested, elicitation accuracy was consistently equivalent or close to that of an expert human advisor. However, we observed a clear failure mode when testing the risk-taking profile, where misunderstandings by the investors and hallucinations within the LLM compound to result in accuracy that is close to random. Overall, we consider this a promising result, as the majority of the time it is effective, and the failure mode observed might be rectified by better context crafting and the addition of contradiction detection; both directions for future research.}

\subsection{RQ2: Effectiveness of personalization} \label{sec:rq2}

\ric{Having shown that automatic preference elicitation is possible, we now examine stage 2 of our study, namely the advisory discussions. Given the inherently personalized nature of financial advice, we expect that the customer preferences obtained during stage 1 will be key to enabling LLM-advisors to provide effective investment advice. Hence, in this section, we compare the performance of an LLM-advisor using only the Baseline Prompt to one that includes the preferences obtained during stage 1 \takehiro{(+Personalized)}. However, as we observed that preference elicitation is not always successful, we also examine what effect elicitation performance has on the LLM-advisor.}

\begin{table}[t]
\centering
\caption{\javier{Investor decision-making effectiveness, expressed as the Spearman's Rho correlation between the investor's asset ranking and the expert asset ranking (higher is better). $^{\dagger}$ indicates statistical improvements (Welch's t-test with $p<0.05$) over the not personalized baseline, while $^{\mathsection}$ indicates significant differences between cases with successful and unsuccessful preference elicitations.}}
\label{tab:ranking_effectiveness}
\resizebox{\linewidth}{!}{%
\begin{tabular}{llccc}
\toprule
\multicolumn{2}{c}{\textbf{\takehiro{Advisor Config}}} & \multicolumn{3}{c}{\textbf{Investor vs.\ Expert (Spearman's Rho)}} \\
\midrule
\multirow{2}{*}{\textbf{\takehiro{Personalization}}} & \multirow{2}{*}{\textbf{Personality}} & \multirow{2}{*}{All} & \multicolumn{2}{c}{Preference Elicitation} \\
\cline{4-5}
& & & Successful & Unsuccessful \\
\midrule
\takehiro{Baseline}
  & None 
  & 0.110 
  & -- 
  & -- \\ 
  +Personalized & None 
  & \textbf{0.310} 
  & \textbf{$0.481^{\dagger\mathsection}$} 
  & -0.228 \\
  \takehiro{+Personalized}
  & +Extroverted 
  & 0.122
  & $0.243^{\mathsection}$
  & -0.286 \\
  \takehiro{+Personalized}
  & +Conscientious 
  & 0.26
  & 0.365
  & \textbf{-0.025} \\
\bottomrule
\end
{tabular}}
\vspace{-5mm}
\end{table}

\subsubsection{Non-personalized Decision-making Effectiveness:}
\looseness -1 \ric{We initially establish how effective the LLM-advisor is without any information regarding the investor. LLM-advisor effectiveness is measured based on how well the investor was able to rank the assets discussed by suitability to them. The primary metric is average Spearman's Rho correlation between the investor ranking and the ground truth ranking (see Section~\ref{ss:stage2}), reported in Table~\ref{tab:ranking_effectiveness} row 1. As we expect, baseline advisory performance is low, with only a very weak positive correlation to the ground truth ranking of 0.11. This indicates that without further evidence, the LLM is not able to meaningfully guide the investor.}

\begin{table*}[ht!]
\centering
\caption{Average participant users' response to advisor assessment questionnaire under different advisor conditions. Columns labeled with advisor condition (Baseline, +Pers., +Cons., +Extr.) contain a 7-point Likert scale (higher is better). 
\emphasize{p} column contains Wilcoxon signed-rank test \emph{p}-values for (RQ2) Baseline vs.\ +Personalized (Pers.), 
and (RQ3) +Conscientious (Cons.) vs.\ +Extroverted (Extr), for both the full data (All) and the subset where the elicitation accuracy is above 0.5. \emphasize{Successful Elicitation} refers to the subset where elicitation accuracy was $\geq 0.5$. \takehiro{For RQ2, this subset consists of pairs for which +Pers elicitation is successful, while for RQ3, it consists of pairs for which both +Extr and +Cons elicitation are successful.}
Boldface indicates significant effects with \(\dagger\) for \(p<0.1\) and \(\ddagger\) for \(p<0.05\).}
\label{tab:survey_results}
\vspace{-2mm}
\resizebox{0.9\linewidth}{!}{%
\begin{tabular}{l
                c c c
                c c c
                c c c
                c c c}
\toprule
& \multicolumn{6}{c}{\textbf{(RQ2) Baseline vs.\ +Personalized}} 
& \multicolumn{6}{c}{\textbf{(RQ3) +Conscientious vs.\ +Extroverted}} \\
\cmidrule(lr){2-7} \cmidrule(lr){8-13}
& \multicolumn{3}{c}{\textbf{All}} 
& \multicolumn{3}{c}{\textbf{Successful Elicitation}}
& \multicolumn{3}{c}{\textbf{All}} 
& \multicolumn{3}{c}{\textbf{Successful Elicitation}} \\
\cmidrule(lr){2-4} \cmidrule(lr){5-7}
\cmidrule(lr){8-10} \cmidrule(lr){11-13}
\textbf{Response Dimension} 
& \textbf{Baseline} & \textbf{+Pers.} & \textbf{p}
& \textbf{Baseline} & \textbf{+Pers.} & \textbf{p}
& \textbf{+Cons.} & \textbf{+Extr.} & \textbf{p}
& \textbf{+Cons.} & \textbf{+Extr.} & \textbf{p} \\
\midrule

\textbf{Perceived Personalization} 
  & 5.759 & 5.724 & 0.838 
  & 5.762 & 5.905 & 0.751
  & 5.500 & 5.500 & 0.663
  & 5.588 & 5.706 & 0.941 \\
  
\textbf{Emotional Trust} 
  & 5.103 & 5.241 & 0.446
  & 5.143 & 5.333 & 0.537
  & 5.038 & 5.154 & 0.600
  & \textbf{4.706} & \textbf{5.235} & \textbf{0.034}$^\ddagger$ \\
  
\textbf{Trust in Competence} 
  & 5.690 & 5.690 & 0.817
  & 5.810 & 5.857 & 0.782
  & 5.962 & 6.077 & 0.538
  & 6.000 & 6.000 & 1.000 \\
  
\textbf{Intention to Use} 
  & 5.310 & 5.483 & 0.505
  & 5.429 & 5.714 & 0.166
  & \textbf{4.885} & \textbf{5.462} & \textbf{0.005}$^\ddagger$
  & \textbf{4.941} & \textbf{5.588} & \textbf{0.013}$^\ddagger$ \\
  
\textbf{Perceived Usefulness} 
  & 5.241 & 5.517 & 0.183
  & 5.381 & 5.810 & 0.194
  & 5.423 & 5.538 & 0.425
  & 5.176 & 5.118 & 0.968 \\
  
\textbf{Overall Satisfaction} 
  & 5.345 & 5.690 & 0.116
  & \textbf{5.429} & \textbf{5.810} & \textbf{0.098}$^\dagger$
  & 5.269 & 5.577 & 0.179
  & 5.118 & 5.529 & 0.244 \\
  
\textbf{Information Provision} 
  & \textbf{5.517} & \textbf{5.966} & \textbf{0.026}$^\ddagger$
  & \textbf{5.714} & \textbf{6.143} & \textbf{0.053}$^\dagger$
  & 5.692 & 5.654 & 0.953
  & 5.588 & 5.765 & 0.490 \\

\bottomrule
\end{tabular}}
\end{table*}

\subsubsection{Personalized Decision-making Effectiveness:}
\ric{Having established our baseline, we now examine the impact that adding the investor preferences collected during stage 1 has, comparing Table~\ref{tab:ranking_effectiveness} row 1 (baseline) to row 2 (personalized). As we anticipated, personalization is beneficial, with investor decision-making effectiveness increasing from 0.11 to 0.31 (average Spearman's Rho correlation to the expert ranking).
However, this correlation is still weak, illustrating that while discussing assets with the LLM-advisor is better than no help at all, our participants are still struggling to evaluate the suitability of financial assets.}  

\ric{\takehiro{This} correlation is an average over all the participants in the user study, regardless of how effective their preference elicitation was in stage 1. Hence, we might ask whether the low correlation is due to the LLM-advisor being confused by poor preference elicitation data. To explore this, Table~\ref{tab:ranking_effectiveness} also reports investor decision-making effectiveness stratified based on whether stage 1 was successful (column 4) or not (column 5).\footnote{We define that an elicitation session is successful if more than 50\% of the investor's preferences were correctly captured}}
\ric{As expected, we see a statistically significant increase in \takehiro{investor decision-making effectiveness when preference elicitation was successful when compared to non-personalized sessions (0.481 vs. 0.110)}. More concerningly, we also see the LLM-advisor has a strong negative influence on the investors' decision-making capability if preference elicitation fails, as illustrated by the negative correlations with the expert in column~5. This result highlights both that effective preference elicitation is crucial, but also that the LLM-advisor can easily influence the investor into making poor decisions, as the human is heavily reliant on the agent to navigate the relatively unfamiliar financial information space.}

\subsubsection{Participant Assessment of the Advisor:}
\ric{So far we have demonstrated that there is a large difference between a \takehiro{non}-personalized LLM-advisor and a personalized one, in terms of how they can alter the decision-making of the investor/participant. But can the participant tell the differences between them?}

\ric{Table~\ref{tab:survey_results} reports the aggregation of the qualitative data we collected from each participant after they finished interacting with each LLM-advisor in terms of 7 dimensions, where we start by focusing on the RQ2-All columns, i.e. comparing the baseline and personalized variants. The important observation to note here is that the participant preference scores for both variants are statistically indistinguishable, except under the quality of information provision criteria. 
This means that our participants cannot tell if the LLM-advisor is personalizing to them, and trust the worse agent just as much as the better one.
Furthermore, if we consider the best case scenario where the preference elicitation was successful (RQ2 Successful Elicitation columns) we observe the same pattern, even though the difference between the baseline and the personalized variants in terms of the effect it has on the participant decision-making is more pronounced. This underlines one of the core risks of using LLM-advisors in the financial domain; since our users are inherently inexpert they lack the fundamental skills to judge to what extent the LLM is providing good advice, meaning that there is no safety net if the LLM makes a mistake. }


\looseness -1 \vspace{2mm}\noindent \ric{\textbf{To answer RQ2}, our results show that a personalized LLM-advisor is able to provide useful financial advice when it has accurate information regarding the preferences of the investor. This is demonstrated by better decision-making capability by participants using the personalized advisor in comparison to the \takehiro{non}-personalized one. However, we also identified two important challenges to adoption. First, the impact the LLM-advisor has is strongly tied to the quality of the preference elicitation data provided, where \takehiro{poor preference elicitation} will cause the agent to actively direct the investor to the wrong assets. Second, while the participants were positive regarding the LLM-advisors across all questionnaire criteria, they were not able to consistently tell the difference between good and bad advisors; leading to an increased risk of humans acting on bad advice.}

\subsection{RQ3: Effectiveness of personalities}
\javier{Once we have confirmed the utility of personalization for LLM-advisors, we now study the effect that the personality of the advisor has on \takehiro{users' financial information-seeking}. As previous studies have shown~\citep{smestad2019chatbot}, chatbot personality can affect the way humans interact with the chatbot, and therefore affect the effectiveness and perception of LLM-advisors. To understand whether personality affects LLM financial advisors, we compare two personalized LLM-advisors on which we have injected a pre-defined personality: an extroverted personality and a conscientious personality.~\footnote{Refer to Section~\ref{ss:ad} for a full description of each personality.} While we could consider the personalized LLM-advisor discussed in Section~\ref{sec:rq2} as a third distinct personality (the base LLM personality of the LLM), we shall not compare it with our personality-injected models, \takehiro{because different sets 
of participants were used in the personalization study and the advisor-persona study.}}

\subsubsection{Decision-making Effectiveness:} \javier{
We first examine the impact of adding personality to the advisors on the decision-making process, by measuring the capacity of the participants to correctly rank the assets (as previously done in Section~\ref{sec:rq2}). As a primary metric, we again use the average Spearman's Rho correlation between the investor ranking and the ground truth ranking reported in Table~\ref{tab:ranking_effectiveness} rows 3 (extroverted advisor) and row 4 (conscientious advisor). }

\javier{We first observe the results for the full set of participants in the user study. Interestingly, we observe a difference between the two advisors, with the conscientious LLM-advisor providing better guidance than the extroverted one (0.26 vs. 0.122). This observation is consistent when we restrict our analysis to those cases where the preference elicitation is successful. While, expectedly, the effectiveness of both advisors improves when the elicitation is successful (0.243 vs. 0.122 in the case of the extroverted advisor and 0.365 vs. 0.26 in the case of the conscientious one), the conscientious advisor has an advantage over the extroverted one (0.365 vs. 0.26).}

\javier{These results highlight that providing different personalities to an LLM-advisor can notably impact the capacity of the advisor to provide useful information to the investors.}

\subsubsection{Participant Assessment of the Advisor:}
\javier{We have observed so far that the use of different personalities affects the user decision-making process. But how do these personalities affect the perception that users have of the LLM-advisor? We observe this in Table~\ref{tab:survey_results}, in terms of the seven dimensions captured during the advisor assessment questionnaire. }

\javier{We first look at the RQ3-All columns, comparing the two personalities. Notably, for the majority of the dimensions, users barely distinguish between both systems. The only answer where we observe a statistically significant difference is the intention to use the system in the future. Surprisingly, despite providing worse guidance to the investor, participants expressed a higher interest in using the extroverted advisor than the conscientious one. When we limit our study to those participants who experienced a successful preference elicitation in both advisor variants, this issue is stressed, as those users also develop a significantly greater emotional trust with the extroverted advisor. }

\javier{These observations are worrisome, as they reveal that the personality of a financial advisor cannot only affect the quality of the advice but also lead the investors to trust more on those systems providing worse advice.}

\subsubsection{Differences in language:} 
\javier{To further understand how personalities affect financial advisory, we analyze the differences in the linguistic patterns provided by extroverted and conscientious advisors. }
\takehiro{Analyzing participants’ reported overall experience from the exit questionnaires in the advisor persona study, over 20\% (7 of 31) described the extroverted advisor as clear, assertive, and cheerful while perceiving the conscientious advisor as straightforward, analytical, yet less confident.\footnote{Participants were unaware of the specific personas during the study.} \javier{Therefore, to} quantify the linguistic differences in the advisors, we conduct a financial sentiment analysis of the utterances generated by each advisor. For each utterance, we count the occurrences of positive, negative, and uncertain words from the Loughran and McDonald Financial Sentiment Dictionary~\cite {loughran2011liability}. We normalize these counts by the length of the sentences and average the results across all dialogues.}

\begin{figure}
    \centering
    \includegraphics[width=0.85\linewidth]{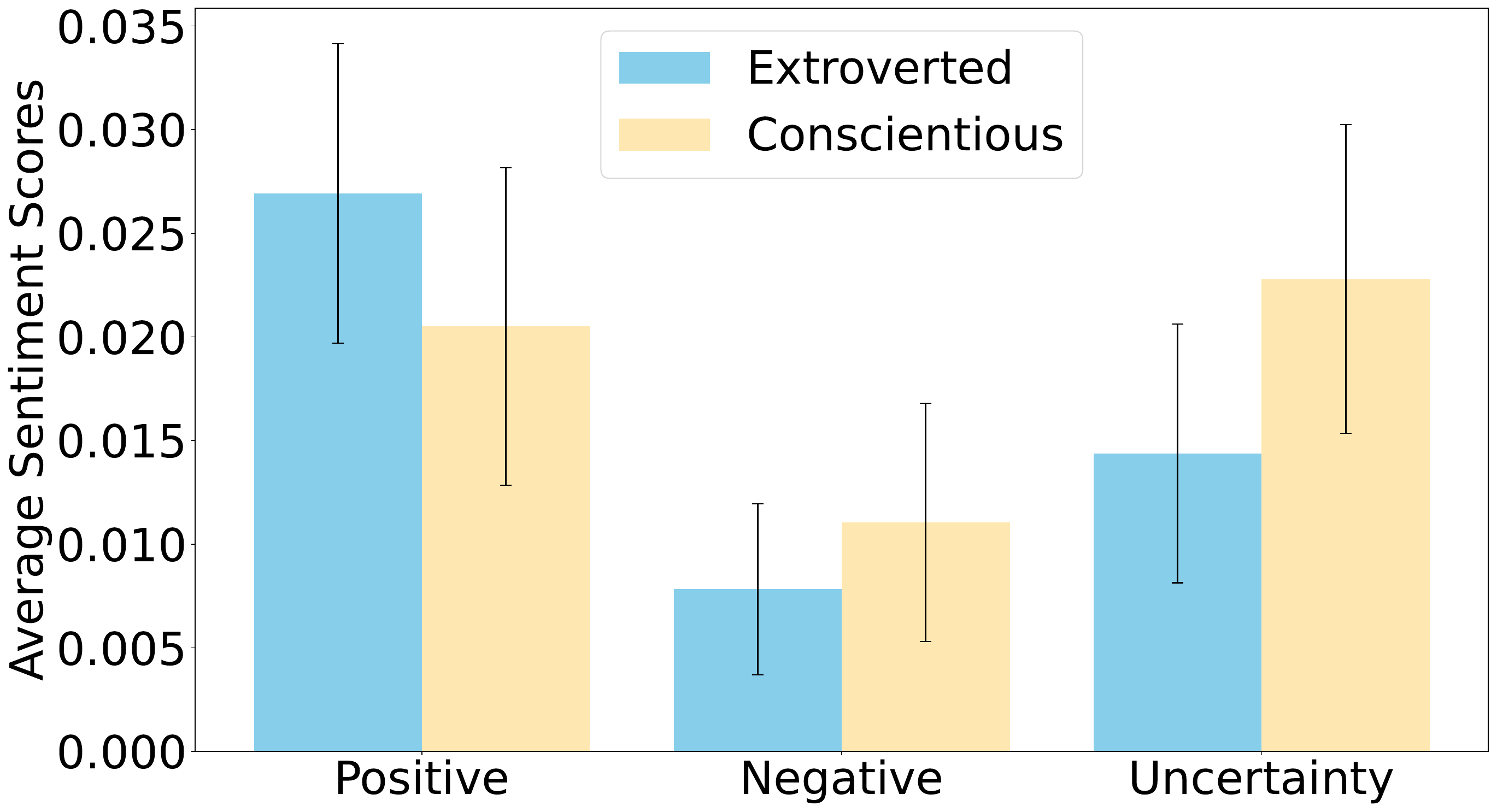}
    \caption{Average sentiment scores by advisor personality (extroverted in light blue and conscientious in pastel orange) and category (Positive, Negative, and Uncertainty). Error bars indicate the standard deviation.
    }
    \label{fig:financial_sentiment_analysis}
    \vspace{-5mm}
\end{figure}

\javier{Figure~\ref{fig:financial_sentiment_analysis} shows the results, showing the extroverted sentiment scores in blue, and the conscientious scores in \takehiro{orange}. For the three sentiment dimensions, differences between advisors are statistically significant (Welch's t-test with $p<0.01$). Figure~\ref{fig:financial_sentiment_analysis} shows that extroverted advisors tend to use more positive language in their interactions, while conscientious advisors prefer negative and uncertain tones. Through manual analysis of the conversation, we observe that this results in the extroverted advisor focusing on the positive aspects of investments while overlooking serious drawbacks, whereas the conscientious advisor provides a more balanced view of the assets. Because of this, participants guided by conscientious advisors \takehiro{may} make more well-informed financial decisions. \takehiro{Meanwhile, the positivity of the extroverted advisor seems more appreciated by the users, which is reflected in higher advisor assessment scores from the post-discussion questionnaire.}}

\vspace{2mm} 
\noindent \javier{\textbf{To answer RQ3, } our results show that different personalities of a personalized LLM-advisor can affect the utility of the provided advice. This is demonstrated by the better decisions of the study participants when using an advisor with a conscientious personality than when using an advisor with an extroverted personality. \takehiro{Moreover}, the personality of the advisor affects the perception of humans towards the system, and it has the risk of leading investors to further trust those systems that provide worse advice.}

\section{Conclusion}
In this paper, we \javier{have conducted a lab-based user study to} examine how \javier{effective} large language models \javier{are} as financial advisors. \javier{We focus} on three core challenges: preference elicitation, investment personalization, and advisor personality.

\javier{First, our analysis shows that LLMs are effective tools for preference elicitation through conversation. In a majority of cases, they are capable of obtaining investor's preferences with an accuracy close to or equivalent to that of an expert human advisor. However, there are some clear failure cases, as LLMs are vulnerable to contradictory statements and hallucinations, \takehiro{which}, in the case of complex investor profiles, can decrease the accuracy of the elicitation to random levels. Although LLMs are promising for elicitation, in a complex domain like finance, investors do not always fully understand their own preferences (or they have difficulties expressing them). Therefore, future work should explore the development of LLM-advisors capable of resolving conflicting user needs.}

\javier{Second, personalizing LLMs to provide investment advice can improve the decisions made by the investors, but only when the personalized LLM-advisor receives accurate information about the investor's preferences. If the preference elicitation is not successful, the agent actively directs the investors to the wrong assets on which to invest. This underscores how crucial a good preference elicitation is for providing useful financial advice.}

\javier{Finally, our results suggest that investors are not necessarily aware of what constitutes good financial advice, and therefore, are vulnerable to acting on bad advice provided by LLMs. In the comparison between a non-personalized and a personalized LLM-advisor, although the personalized system led to better decisions, participants were unable to distinguish between the systems. More worryingly, when comparing two personalized advisors with extroverted and conscientious personalities, we observed that, even though the extroverted advisor provided lower-quality advice, participants trusted this advisor more than the conscientious one.}

\javier{Our findings highlight that, while personalized LLM-advisors represent a promising research direction, their use in high-stakes domains like finance is not free of risks: due to the limitations of LLMs at capturing complex investment preferences, and the difficulty of investors to discern whether the advice they receive truly serves their interests, LLMs have a notable risk to drive investors to bad financial assets (leading not only to a low satisfaction but also to potentially large monetary losses).} \javier{However, these drawbacks open interesting research directions not only from a system perspective, but also from a human-centered approach: automated advisory development where we do not just focus on improving the quality of automated systems to guide investors, but also on how the investors will adopt, trust and interact with these AI agents~\citep{chiou2023trusting,human_trust}.}

\bibliographystyle{ACM-Reference-Format}
\bibliography{sigir2025}








\end{document}